\documentclass[conference]{IEEEtran}
\IEEEoverridecommandlockouts
\usepackage{cite}
\usepackage{amsmath,amssymb,amsfonts}
\usepackage{algorithmic}
\usepackage{graphicx}
\usepackage{textcomp}
\usepackage{xcolor}

\def\BibTeX{{\rm B\kern-.05em{\sc i\kern-.025em b}\kern-.08em
    T\kern-.1667em\lower.7ex\hbox{E}\kern-.125emX}}
\begin{document}
\title{Autoencoder-based Condition Monitoring and Anomaly Detection Method for Rotating Machines\\
}

\author{\IEEEauthorblockN{Sabtain Ahmad,
Kevin Styp-Rekowski,
Sasho Nedelkoski,
Odej Kao}
\IEEEauthorblockA{Distributed and Operating Systems\\
TU Berlin\\
Berlin, Germany\\
s.ahmad@campus.tu-berlin.de,~\{firstname\}.\{lastname\}@tu-berlin.de}}

\maketitle

\begin{abstract}
Rotating machines like engines, pumps, or turbines are ubiquitous in modern day societies. Their mechanical parts such as electrical engines, rotors, or bearings are the major components and any failure in them may result in their total shutdown. Anomaly detection in such critical systems is very important to monitor the system's health. As the requirement to obtain a dataset from rotating machines where all possible faults are explicitly labeled is difficult to satisfy, we propose a method that focuses on the normal behavior of the machine instead. We propose an autoencoder model-based method for condition monitoring of rotating machines by using an anomaly detection approach. The method learns the characteristics of a rotating machine using the normal vibration signals to model the healthy state of the machine. A threshold-based approach is then applied to the reconstruction error of unseen data, thus enabling the detection of unseen anomalies. The proposed method can directly extract the salient features from raw vibration signals and eliminate the need for manually engineered features. We demonstrate the effectiveness of the proposed method by employing two rotating machine datasets and the quality of the automatically learned features is compared with a set of handcrafted features by training an Isolation Forest model on either of these two sets.  Experimental results on two real-world datasets indicate that our proposed solution gives promising results, achieving an average F1-score of 99.6\%.
\end{abstract}


\section{Introduction}
In the modern day industry, machine systems are becoming more complex and fulfill critical tasks. To enhance their reliability, the condition of the system should be monitored. Any rotating machine, e.g. a pump, compressor, or steam turbine, will eventually reach a point of poor health. One effective strategy for enhancing their reliability and cost-effective maintenance is to utilize Condition Monitoring (CM) and Prognostics and Health Management (PHM). The aim is to identify unexpected anomalies, faults, and failures \cite{orchard2009particle}.

Prognostics systems built on a data-driven approach acquire data in-situ using a network of sensors that monitor the system \cite{nectoux2012pronostia}. The dataset evolving from the measurements of the sensors usually has a high dimensionality and may also contain unwanted interference and noise. This dimensionality of the data has a direct impact on the training time as well as the accuracy of neural network-based models. A workaround for dealing with such a problem is to reduce the dimensionality of the input signal by extracting features that carry the health information of the system. Several methods have been proposed for achieving this, for instance, extracting temporal features by computing the root mean square (RMS), Skewness, Kurtosis, or Peak to Peak distance \cite{tsay2000outliers}.

The main attributes of an ideal anomaly detection and condition monitoring system include the ability to collect useful features and the utilization of these features to identify the deteriorating condition of the machine by observing the deviation from the normal (healthy) behavior. Manual feature engineering methods require domain knowledge combined with “trial and error” strategies. However, the advent of deep learning and recent progress in autoencoder based models has provided an alternative way for feature extraction and dimensionality reduction. By stacking up layers to form deep autoencoders and by reducing the number of units in the hidden layers, it is expected that hidden units will extract features that will represent the data. The best features for the task at hand are learned directly from the data, thus avoiding ad-hoc trial and error strategies. Since the autoencoder creates a reduced representation of the data, it seems intuitive to use this representation for anomaly detection. The assumption is that the autoencoder only learns to map normal data points or inliers and does not include anomalies in the trained representation. Hence, trying to reconstruct anomalous data points will fail and incur huge losses. Finally, this reconstruction error can be used as a basis for calculating anomaly scores and labeling of unseen data points.

For rotating machine (RM) anomaly detection in general, it is much easier to collect large amounts of data than to accurately obtain their corresponding labels. Particularly in cases where faults or degradation evolve naturally over time. Correctly assigning labels is susceptible to the data ambiguity issue, especially at a pivotal stage when the machines exhibit early signs of failure but are far from obvious when equated with the fully developed faults. Moreover, anomalies are rare events, having prior knowledge about all possible anomalies is almost impossible, so the attention is shifted from anomalous to normal states for which data is available in large quantities. The motivation is that by modeling the normal behavior of the machine the system will also be able to detect anomalies that have not been observed previously.

\begin{figure}[t]
\begin{center}
\includegraphics[scale=0.8]{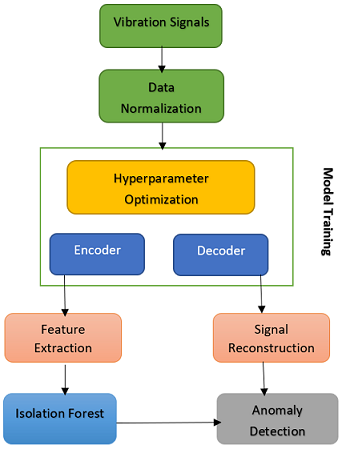}
\end{center}
\caption{Framework flowchart of the proposed method.} 
\label{fig:flowchart}
\end{figure}

We propose a framework for RM condition monitoring and anomaly detection based on Long short-term memory (LSTM) autoencoder networks. We demonstrate its applicability in both anomaly detection and condition monitoring by evaluating it on two real-world datasets. The advantages of the autoencoder based approach include the ability to work without any preprocessing, without any predetermined transformations such as FFT, without any manual feature engineering, without any feature selection, and the fact that it does not limit itself to the preidentified anomalies; it has the potential to detect new anomalies which have never been seen before. We also show that the time-domain features learned in the process can be used to enhance the performance of a simpler detection model such as Isolation Forest and thus can obviate the need for manual feature extraction. The flowchart of the proposed method is displayed in Fig.~\ref{fig:flowchart}. The measured vibration signals will be preprocessed and afterward, the training of the model is completed that is able to extract features and reconstruct the signal for the final anomaly detection.

The remainder of this paper is organized as follows. We will discuss related works in Section \ref{rel_work}. Section \ref{framework} discusses our proposed LSTM-based autoencoder method for robust feature extraction and anomaly detection. Sections \ref{eval} and \ref{results} discuss the experimental results, while the conclusions and future works are presented in Section \ref{conclusion}. 
 
\section{Related Work}
\label{rel_work}
A plethora of studies has been conducted in the field of RM fault/anomaly detection and time-series anomaly detection in general \cite{aggarwal2017outlier}. Features and their generation are important concepts in data analysis and anomaly detection. The use and selection of features are crucial for measuring differences in data thus detecting anomalies. Traditionally, features extracted from time or frequency domain have been used for monitoring the condition or computing the remaining useful life (RUL) of the machine \cite{qiu2006wavelet, tobon2012data}. Many extracted features are influenced by operating conditions and are insensitive to anomalies. Researchers have tried to increase the performance of anomaly detection methods for RMs by performing multivariate analysis while using multiple features together,  \cite{xia2012spectral} computed 21 bearing features using signal processing techniques. Others have tried to achieve this by constructing features for specific failure modes and performing multivariate analysis based on them as did the authors in \cite{tian2015motor} by constructing 5 features each of which was sensitive to a different failure mode. However, such manually extracted features are not generalizable and fail to provide useful machine health information in especially unknown cases.


More recently, \cite{ali2015application} tried to classify various bearing classes by using a set of time-frequency domain features and artificial neural networks. \cite{ince2016real} replaced the manual feature extraction step by applying 1-D convolutional neural networks (CNN) to raw motor signals. The evaluation performed on bearing fault detection demonstrated the superiority of their approach compared to conventional feature extraction methods. \cite{sun2017intelligent} trained an autoencoder for feature extraction and used these features to train a supervised fault detection classification model. \cite{razavi2017one} studied several one-class classifiers such as nearest neighbors and k-means for detecting faulty rotor bars in an induction motor. They concluded that the k-nearest neighbor method stood out among all the tested methods.

Deep neural network-based architectures in particular autoencoders are successfully employed for supervised classification of faults into different fault categories by using time or frequency domain features extracted using prior knowledge \cite{jia2016deep}. A probabilistic framework for anomaly detection in natural gas consumption time series is introduced in \cite{akouemo2016probabilistic}. However, the prediction method predicts the consumption levels using other independent variables and does not incorporate the temporal information that the data had to offer. \cite{martinelli2004electric} employed an vanilla autoencoder to detect anomalies in the electric power system by embedding the temporal information using sliding windows. Reconstruction errors obtained on sliding windows were used to compute anomaly scores. The Inclusion of temporal information using a sliding window works well in some cases but is not scalable usually.  

The current RM anomaly detection methods, as discussed above, face one or several of the following limitations: labor-intensive manual feature extraction, the requirement of accurately labeled datasets, or failure to incorporate temporal information. Time series data such as sensor data is best modeled as a sequence where the data point at each timestep is dependent on the previous data points. LSTM-based autoencoders are capable of dealing with the time-series sequences and can take variable length input. 
To overcome issues described above, in this study, we train an LSTM-based autoencoder over the vibration signals to autonomously monitor the condition of RMs and extract time-domain features to provide an alternative for manual feature extraction. To the best of our knowledge, this is the first time that an unsupervised method based on LSTM-autoencoders
is used for identifying fault/anomalies in accelerometer vibration signals.

\section{Rotating Machine Anomaly Detection}
\label{framework}

We address the problem of anomaly detection in abnormal vibratory phenomena captured through the accelerometer sensors mounted on RMs which should indicate a deterioration of the system. As opposed to traditional health prognostic systems that usually encapsulate feature extraction and anomaly detection as distinct blocks, the proposed system takes directly raw time-series vibration signals as input and it can efficiently learn optimal features and based on these features determine the system's health.

In the case of multi-dimensional input, the set of M sensors $\{m_1, . . . , m_M\}$ (also called generators) are used to capture the behavior of a RM. This measured data is fed as input to an LSTM-based autoencoder (LSTM-AE) that is trained over the vibration signals via batch gradient descent to minimize a reconstruction error term between an original signal and its reconstruction. Specifically, the encoder maps an input vector \textbf{\textit{x}} to a lower-dimensional hidden representation \textbf{\textit{h}} by an affine mapping following a nonlinearity and the decoder correspondingly generates an estimation ${\textbf{\textit{x’}}}$ of the input vector ${\textbf{\textit{x}}}$.

AEs belong to the unsupervised representation learning class which try to model the data distribution through the discovery of a set of latent representations, also called embeddings, whose variations capture most of the structure of the original data distribution \cite{alain2014regularized}. These hidden-layer units or low-dimensional embeddings force the AE model to learn the key representations from the original vibration signal. The encoder generates a rich non-linear set of features from the sensory data and the decoder learns to reconstruct the original signal using these features. The motivation to use autoencoders is their ability to detect anomalies based on the fact that anomalies are quite rare and deviate greatly from the general pattern in normal healthy data. The model is trained with the aim to learn the normal behavior of a RM, thus not recognizing anomalies.



\begin{figure*}[t]
\begin{center}
\includegraphics[scale=0.9]{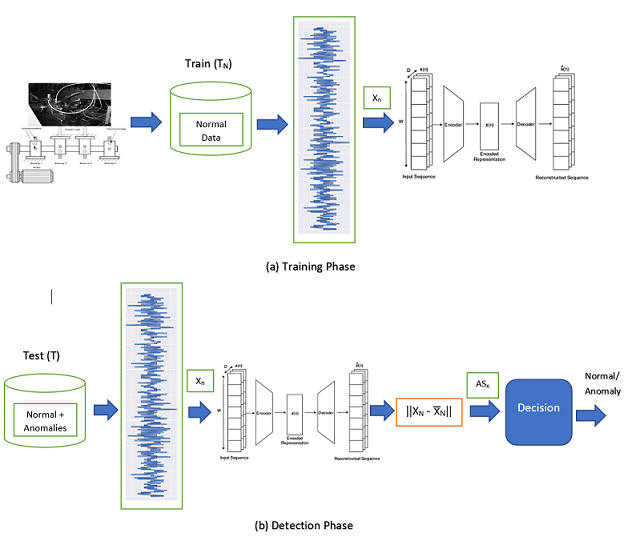}
\end{center}
\caption{Condition monitoring of RMs through diagnosis of anomalies using LSTM-AE.} 
\label{fig:lsmt_ae}
\end{figure*}

The encoder and decoder are two main components of the network and both are based on Long short-term memory network (LSTM) units.
LSTMs are a type of recurrent neural network (RNN) that can integrate the temporal information into the network and maintain a hidden state vector which acts as a memory for the past information \cite{greff2016lstm}.

\paragraph{Encoder}
We observe the input data sequence denoted by ${\mathbf{X = (x^{(1)}, x^{(2)}, ..., x^{(N)} )}}$, where ${\mathbf{x^{N} = (x_1^{(N)}, x_2^{(N)}, ..., x_T^{(N)} )}}$ 
${\in {R^{Txd}}}$, meaning that for each index N there is a d-dimensional time-series sequence with T timesteps each. We use the RNN to process the variable input sequence and to extract the sequential information from the time-series data. For each sequence this is given by;

\begin{equation}
C_t' = tanh(W_C . x_t + R_C h_{t-1} + b_C) 
\end{equation}

where ${C_t}'$, $x_t$ and $h_{t-1}$ are memory state, input and output vectors from last step and $W_C$, $R_C$ and $b_C$ are input weights, recurrent weights and the bias. Tanh is used as a non-linear activation function, whose output range lies in the interval $[-1,1]$.

The input sequences are passed through the encoder part of the LSTM network which encodes an input sequence or batch of sequences using LSTM units and updates its hidden state. The output of the encoder is given by;

\begin{equation}
h_t = \sigma^{e}_\phi(x_t , h_{t-1}) 
\end{equation}

where $h_t$ is the output of the $i^{th}$ LSTM-encoder, $\phi$ represents the parameter set of the encoder and to avoid the vanishing and exploding gradients issue, $\sigma$ for both encoder and decoder is chosen as ReLU activation function.


\paragraph{Decoder:}
The representations obtained from mapping the vibration signal to lower-dimensional embeddings through the encoder are used by the decoder to reproduce the original signal. The Output from the last encoder of the network becomes the input to the LSTM-decoder network:

\begin{equation}
h_t{'} = \sigma^{d}_\varphi(h_t , h'_{t-1}); \; \; \; x'_t = \sigma(h'_t) 
\end{equation}

where the set of parameters of the decoder is represented by $\varphi$ and $x'$ is the reconstructed input which is used to compute the \textit{reconstruction error (RE) = ${|| x_t - x'_t ||^2}$}, also known as the Mean Squared Error. The error is required to update the network's encoder and decoder parameters and later, compute the anomaly scores.

The AE model should be sensitive enough to reproduce the original signal but insensitive enough to the training data and noise, such that the model learns a generalizable representation of the data. During training, the focus of the model lies in learning the normal behavior of the machine, hence anomalies are not included in the training data. Consequently, during prediction, a slight deviation from the normal behavior would increase the \textit{reconstruction error (RE)}. So, monitoring the increase in reconstruction error gives the possibility to detect anomalies but also anticipate the fault in advance by identifying the degradation point, the timestamp after which the RE starts to increase.

\subsection{Model Training}
The raw vibration data is normalized first to have zero mean and unit variance. We divide the RM's vibration data into four sets: a training set ($T_N$), two validation sets ($V_N$ \& $V_A$), and a test set ($T_A$). The distribution of data among these four sets is as follows: $T_N$ 70\%, $V_N$ \& $V_A$ 5\% (each) and $T_A$ 20\%. $T_N$ consists only of normal sequences and is used for training the LSTM-AE model. NASA bearing dataset (dataset-1, details in section \ref{eval} ) does not provide explicit labels for each sequence, so we assume initial 70\% of the data to be normal as machines have a low probability of being in a faulty state, meaning that we assume anomalies to occur rarely compared to the normal data. According to the ground truth in the provided datasets, failures occurred only at the end of each run-to-failure experiment and thus the first 70\% is used for the training of the model. The architecture of the LSTM-AE is designed as such to allow the model to have enough capability for feature learning and secondly, the number of units of the next hidden layer is set smaller than that of the previous layer so that feature learning can be viewed as a signal compression process, as shown in Fig.~\ref{fig:lsmt_ae} (a). Here, the training process with the train set $T_n$ is depicted, together with the compression process of the raw signal within the autoencoder. The weights of the LSTM-AE are updated via stochastic gradient descent using mini-batches and the Adam \cite{duchi2011adaptive} optimizer is utilized to speed up the training process. Batch normalization allows faster and stable training of deep neural networks. To restrict the model from overfitting, dropout is used and to avoid exploding gradients, gradient clipping is applied. Model hyperparameters are learned using the Bayesian optimization method, including the size of mini-batches, learning rate, and weight decay.

\subsection{Anomaly Monitoring Process}

The detection process of anomalies consists of converting reconstruction errors into anomaly scores(AS) for each input sequence and using these scores to obtain a threshold, characterizing the normal behavior of the machine.

After the model has been trained, sequences in $V_N$ are passed through the model to get the reconstruction errors which in turn are used to estimate the parameters of a \textit{Normal distribution} ($\mu$ $\Sigma$) using maximum likelihood estimation (MLE), similar to  \cite{malhotra2016lstm}. The probability $p_i$ of obtaining the reconstruction error $e_i$ is given by the value of the Normal distribution at $e_i$.
Using the $\mu$ \& $\Sigma$, the anomaly score for a datapoint ${\mathbf {x_t^{(N)}}}$ is computed as follows:

\begin{equation}
a_i = (e_i - \mu)^T \Sigma^{-1} (e_i - \mu) \label{eq:3.3} 
\end{equation}

where $a_i$ is the desired anomaly score, $e_i$ is RE obtained for a sequence ${x_t^{(i)}}$ and $\mu$ $\Sigma$ are the mean and variance of a multivariate Gaussian distribution. 


During the initialization phase, an anomaly score threshold $\tau$ is also learned using a validation set $V_A$ that may contain examples from anomalous sequences alongside with the normal sequences. Unseen sequences within $T_A$ are classified as follows: if a sequence has an anomaly score $>$ $\tau$ it will be labeled as an anomaly, otherwise as normal. This is also depicted in Fig.~\ref{fig:lsmt_ae} (b) where the raw measured signal is input to the trained autoencoder model that generates reconstruction errors and anomaly scores. These are then used within the described decision process to determine whether the signal was normal or anomalous.

The assumption here is, that as the monitored equipment degrades or faces a failure, this disrupts the normal working of the machine and affects the interaction between different variables which can be measured by accelerometers, especially in form of vibrations. As the sensor values start deviating from the normal working condition of the machine, it is expected to see an increased error in the reconstruction of the input. By monitoring the reconstruction error and the anomaly score, an indication of the health of the monitored machine can be derived.

\subsection{Feature Extraction}
Feature learning is a critical step in improving the performance of anomaly detection models due to the multidimensionality of data that is input into the model. In general, machine vibration signals comprise a stationary vibration part, a random vibration part, and noise \cite{singh2004vibration}. We study two methods for the inspection of vibration signals: automated feature extraction based on LSTM-AE and manual feature extraction based on classic signal processing methods. Isolation Forest is used to evaluate the effectiveness of these two feature extraction methods.

Isolation Forest is a random forest-based anomaly detection algorithm that utilizes isolation to determine anomalies in data \cite{liu2008isolation}. The Isolation Forest (IF) algorithm is based on Decision Forests, an ensemble method that uses the averages of outputs from many different trees.

\subsubsection{LSTM-AE based Feature Extraction}
One of the main characteristics of an autoencoder that is more powerful for finding intrinsic data structures by reducing data dimensionality through non-linear transformations than Principal Component Analysis (PCA) \cite{manning2018pca}. 

The encoder from the trained LSTM-AE is used to perform feature extraction. The process is done by reducing the number of units in the hidden layer, it is expected that the hidden units in the encoder network will extract features that will represent the data. To learn more abstract features, multiple AEs are stacked together to form a stacked AE, in which the output of each hidden layer is connected to the input of successive hidden layers. A stacked AE applies dimensionality reduction in a hierarchical manner, obtaining more abstract features in higher hidden layers which lead to a better reconstruction of the data \cite{vincent2010stacked}.


\subsubsection{Manual Feature Extraction}
Following time-domain statistical features are generally used to detect incipient machine faults/anomalies: mean ($\mu_x$), root mean square (RMS), percentiles (25th, 50th \& 75th ), max absolute value, standard deviation ($\sigma_x$),peek-to-peek, skewness, kurtosis, entropy, and AR-coefficients. The last five features are described through the following equations:

\begin{equation}
Peak{-}peak (p{-}p) = abs(Max(x)) + abs(Min(x))
\end{equation}

\begin{equation}
Skewness = \frac{\sum_{i=1}^{N}(x_i - \mu_x)^3}{N \sigma_x^3}
\end{equation}

\begin{equation}
Kurtosis = \frac{\sum_{i=1}^{N}(x_i - \mu_x)^4}{N \sigma_x^4}
\end{equation}

\begin{equation}
Entropy = \sum_{i}p_i \log p_i
\end{equation}

\begin{equation}
AR{-}coefficients = \sum_{k=1}^{p} a_k x[n-k] + e[n]
\label{eq:ar}
\end{equation}

where in equation \ref{eq:ar};  p is the degree of the AR model, x[n] is a signal composed of b data points, $a_k$ is real-values AR coefficient and e[n] is white noise.

\section{Evaluation}
\label{eval}

There are two datasets considered for the evaluation of the proposed method. We evaluate and test the applicability of the proposed method in Condition Monitoring using the IMS bearing dataset \cite{lee2009rexnord}(Dataset-1). In addition, the effectiveness to find anomalies in unsupervised settings is evaluated using a private industry dataset from vibration data of RMs. Evaluations performed on these two datasets demonstrate the ability of the method to detect anomalies within the industry dataset and perform condition monitoring on the bearing dataset, as the IMS bearing dataset doesn’t provide explicit labels but instead contains a degrading health state scenario of the bearing under experiment. Below, we introduce the datasets and illustrate the performance of the proposed method on these datasets. 

\subsection{Dataset-1}
This dataset was gathered from a run-to-failure experimental setting, involving four bearings and is subdivided into three datasets, each of which consists of the vibration signals from these four bearings \cite{lee2009rexnord}. These sets were collected from three test-to-failure experiments which were performed independently, and failures occurred at the end of the test. Table~\ref{tab:rm_nasa} represents the properties of the collected data from these experiments.

\begin{table}[t]
\caption{Dataset-1: NASA Bearing dataset description}
\begin{center}
\begin{tabular}{|c|c|c|c|}
\hline
\textbf{Set \#} & \textbf{Batches}& \textbf{Batch Size}& \textbf{Anomaly} \\
\hline
  Set1 & 2156 & 4 x 20480 & B3 and B4 \\
  \hline
         Set2 &  984 & 4 x 20480 & B1 \\
         \hline
         Set3 &  6324 & 4 x 20480 & B3\\
        \hline

\end{tabular}
\label{tab:rm_nasa}
\end{center}
\end{table}

\begin{table}[t]
\caption{Dataset-2: Industry dataset description}
\begin{center}
\begin{tabular}{|c|c|c|c|}
\hline
\textbf{Machine} & \textbf{Batches}& \textbf{Batch Size}& \textbf{Anomalies} \\
\hline
 RM-1 & 1176 & 6144 & 53 \\ 
 \hline
        RM-2  & 1463 & 2401 & 4 \\ 
        \hline
        RM-3  & 2204 & 2401 & 4 \\ 
        \hline
        RM-4 & 1452 & 2401 & 3 \\ 
        \hline
        RM-5 & 1452 & 2401 & 4 \\
        \hline

\end{tabular}
\label{tab:rm_2}
\end{center}
\end{table}

\subsection{Dataset-2}
\begin{figure}[t]
\centering
\includegraphics[width=1.0\linewidth]{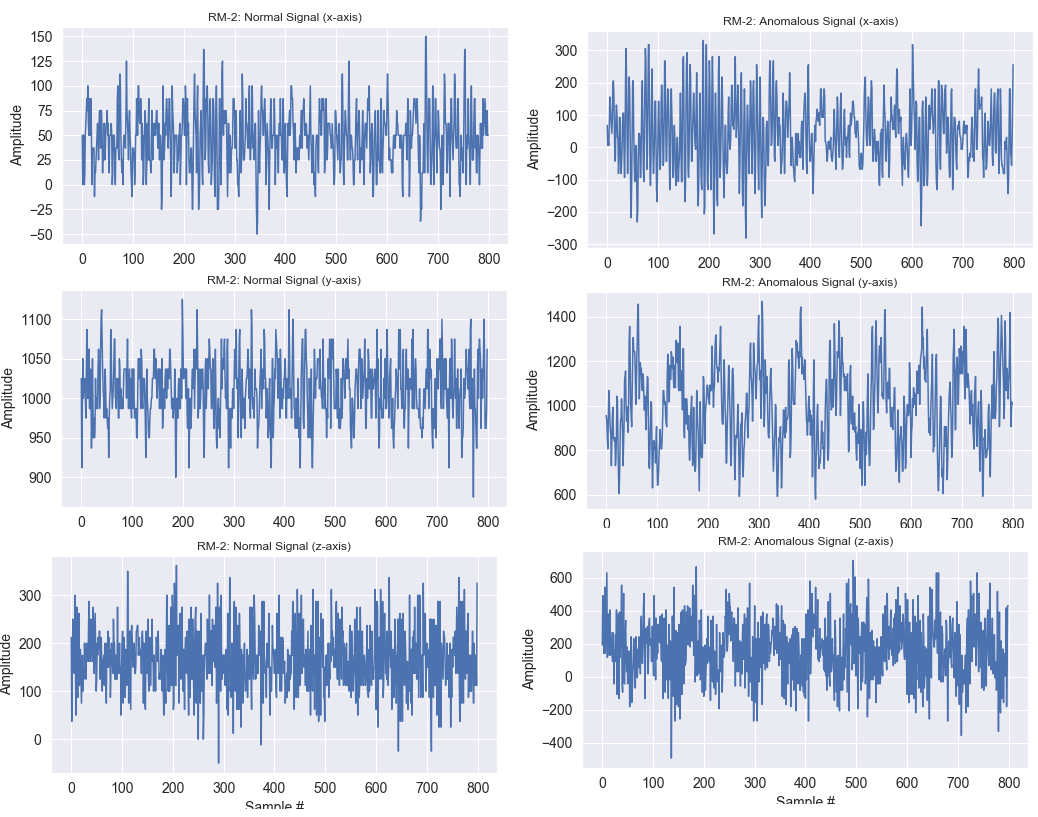} 
\caption{Example of a normal (Left) and anomalous (Right) signal from RM-1, rows corresponding to the x, y and z-axis of the accelerometer.} \label{fig:p2_ex}
\end{figure}

This dataset consists of data from five different RMs (RM-1 to RM-5) which are of the same kind but build-wise unique. Data from each RM contains 3-dimensional vibration signals measured over time, captured using the accelerometer sensors attached to the housing of the RM. The signal recordings were taken in batches and the batch sizes vary, depending upon the RM. Considering the condition of the machine at the time of recording, labels are assigned to each batch, 0 representing normal and 1 anomalous, respectively. Table~\ref{tab:rm_2} presents the summarized statistics of the industry dataset. The difference between normal and anomalous signals is shown in Fig.~\ref{fig:p2_ex}, vibrations in all three directions (x, y, z) differ significantly for anomalous signals compared to normal vibration signals. These differences can have a variety of characteristics as the amplitude of different frequencies differs between the two signals.

\section{Experiments and Results}
\label{results}

We demonstrate our method’s applicability in condition monitoring \& anomaly detection with three general settings where in the first one, the autoencoder is trained with 70\% of the data, 10\% (split between two validation sets) is used to calculate the threshold, and remaining 20\% of it is kept for evaluation to determine the prediction ability of the proposed approach for an anomaly detection task. In the second experimental setting, each run-to-failure experiment is simulated by training the LSTM-AE model with the available data from the non-failing RMs to monitor the condition of the machine in which failure occurred during the run-to-failure experiment. The trained model is used to monitor the status of the faulty machines to demonstrate that once the model is trained, it is capable of being applied to different RMs of a similar kind, enabling the transfer of rare anomaly knowledge from RMs to other RMs of a similar kind.

The third setting consists of a method for performing automated feature extraction from vibration signals. Features extracted using this setting are compared against a set of handcrafted features by training Isolation Forest on these two features sets separately.

\subsection{Setting 1: Online Prognostic}
In this setting, as discussed, the autoencoder learns to reconstruct the normal behavior of an RM using only 70\% of the available healthy/normal data in order to evaluate the prediction performance of the model in an online monitoring phase. The threshold $\tau$ is obtained using a validation set of size 5\% to classify the samples as normal/healthy or anomalous. Anomaly scores for each data sample are calculated using the reconstruction errors and data points with anomaly scores larger than $\tau$ are labeled as anomalies. The point at which the anomaly score crosses the threshold and starts to increase gradually is considered as the degradation point.

\begin{figure}[t]
\centerline{\includegraphics[scale=0.7]{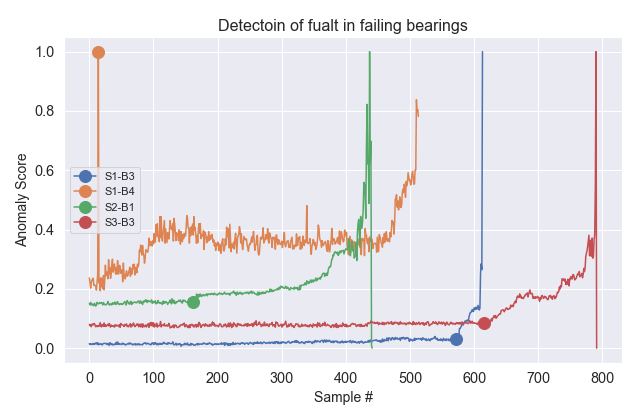}}
\caption{Deteriorating condition of the failing bearings captured by the model. High anomaly scores indicate that  system is more likely to be in faulty state. Circles mark the start of the degradation for each bearing.} 
\label{fig:beairng_fault}
\end{figure}

\paragraph{Result: Dataset-1}
Here we study 4 cases of failing bearings (B1 to B4) of the different sub-datasets (S1 to S3), S1-B3, S1-B4, S2-B1, and S3-B3, for early fault detection. The model trained on individual bearings is able to predict the degradation point and capture the propagation of fault in the simulated run-to-failure experiment as is depicted in Fig.~\ref{fig:beairng_fault}. As the fault in every failing bearing occurs at the end, Fig.~\ref{fig:beairng_fault} displays the anomaly scores of the last 620, 450, and 780 batch samples for S1-B3 \& S1-B4, S2-B1, and S3-B3, respectively. For every failing bearing in each dataset, the proposed method generates a trend corresponding to the health status of the bearing based on the vibration sensory data. As the bearing health starts to degrade, the anomaly score tends to increase and once it passes the degradation point (marked with the filled circles) it increases gradually. By following this trend and abrupt changes in the scores one can feasibly detect the failure. The degradation point for each bearing was calculated by identifying the abrupt change and continuous increasing trend in anomaly scores. Sample no. 2039, 1704, 667, and 5241 were identified as degradation points for S1-B3, S1-B4, S2-B1, and S3-B3, respectively. The circle on the score line indicates the degradation starting point.

\begin{figure}[t]
\centerline{\includegraphics[scale=0.5]{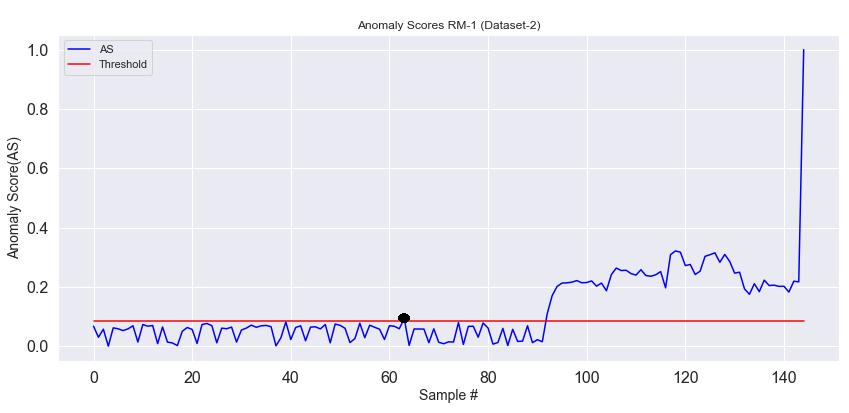}}
\caption{Anomaly scores for RM-1 from the second dataset: Red straight line indicates the threshold, data points above this red line are labeled as anomalous, otherwise normal. Black circles indicate false positives.} \label{fig:p1_result}
\end{figure}

\begin{table}[t]
\caption{Setting 1 results: individual RM anomaly detection scores.}
\begin{center}
\begin{tabular}{|c|c|c|c|c|c|}
\hline
\textbf{Machine} & \textbf{Precision}& \textbf{Recall}& \textbf{TPR} & \textbf{FPR} & \textbf{F1-score}  \\
\hline
         RM-1 & 0.993 & 0.993 &  0.981 & 0.010 & 0.993 \\ \hline
         RM-2 & 1.0 & 1.0 &  1.0 & 0.0 & 1.0 \\ 
         \hline
         RM-3 & 1.0 & 1.0 &  1.0 & 0.0 & 1.0 \\  
         \hline
         RM-4 & 0.994 & 0.991 &  1.0 & 0.007 & 0.993 \\  
         \hline
         RM-5 & 0.965 & 0.972 &  0.25 & 0.007 & 0.967 \\
        \hline

\end{tabular}
\label{tab:rm_2_result}
\end{center}
\end{table}

\paragraph{Result: Dataset-2}
The second dataset consists of 5 different RMs with anomalies, for each machine there is a different number of anomalies (see Table~\ref{tab:rm_2}). To detect anomalies from each individual machine, we conduct five different experiments (one for each machine) using only the non-anomalous (normal) data to train the autoencoder network. The anomaly detection results are evaluated using, Precision, Recall, True positive rate (TPR), False positive rate (FPR), and F1-score. These evaluation metrics are weighted per class.

As described, model performance is evaluated over the test set consisting of 20\% of the data. For instance, the test set for RM-1 consisted of 145 (92 normal, 53 anomalous) sequences in total. Each sample from the test set was reconstructed using the trained autoencoder network and was labeled as anomalous if its anomaly score was larger than the threshold, normal otherwise. The trained model was able to detect 52/53 anomalies without raising many false alarms (just one false positive). Fig.~\ref{fig:p1_result} displays the anomaly scores for each data point; sequences below the threshold (marked by the red line) are classified as normal and data points with anomaly scores larger than the threshold are labeled as anomalies. Evaluation scores (precision, recall, TPR, FPR, F1-score) for individual RMs from the second dataset, averaged over 10 repetitions of the experiments are presented in Table~\ref{tab:rm_2_result} in which the model performs very well with F1-scores close to 1 and is able to detect anomalies without generating many false positives which is indicated by the high TPR and low FPR. Based on the results, the LSTM-AE can effectively extract discriminative features directly from the raw vibration data and achieve a competitive anomaly detection rate. For the test data, an overall detection F1-score of 99\% and TPR over 93\% for each machine is obtained.

\subsection{Setting 2: Condition Monitoring}
In dataset-1, during each run-to-failure experiment, only one out of 4 bearings faced failure while the other 3 remained in healthy condition except for experiment-1 in which a fault occurred in two bearings at the end. The question of interest is whether the knowledge extracted from the healthy bearings can be applied to detect the deteriorating condition of a faulty bearing and raise alarm well before the total failure actually happens. To validate this, in this setting only the data from the healthy bearings is used to train the model and this trained model is then used to evaluate the condition of the previously unseen faulty bearings.

\begin{figure*}[t]
\centerline{\includegraphics[scale=0.8]{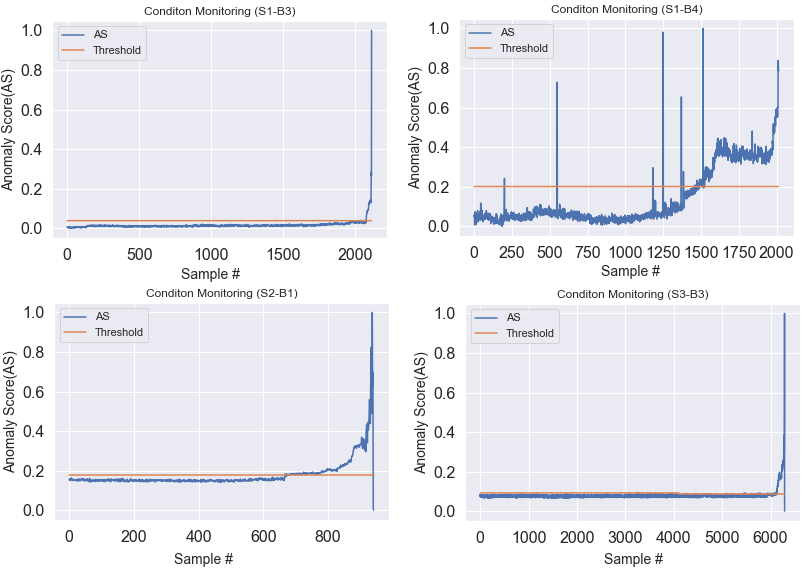}}
\caption{Anomaly scores for the condition monitoring of the faulty bearings by modeling the normal working behavior of the remaining bearings using the other non-failing bearings from the run-to-failure experiments.} 
\label{fig:cm_ims}
\end{figure*}

Three simulations, one per sub-dataset are performed as follows: The \textbf{Set1} model is trained using only training data from the non-failing bearings S1-B1 \& S1-B2 with the aim to learn the functioning of healthy bearing and apply this learned knowledge to monitor and detect the deteriorating condition of the failing bearings S1-B3 \& S1-B4. Similarly, for \textbf{Set2}, the training data comprised of S2-B2, S2-B3, and S2-B4 and while the test set consisted of S2-B1 and for \textbf{Set3}, training data consists of S3-B1, S3-B2, S3-B4 and test set of S3-B3. Fig.~\ref{fig:cm_ims} displays the output anomaly scores of the four failing bearings from these experiments. The trend it generates for each bearing is in accordance with the ground truth, from the beginning until near the end the bearings remain in healthy condition, thus low anomaly scores are correctly calculated. The fault starts to appear only at the very end, which is captured by the model by a continuous increase in the anomaly scores.

\begin{figure}[h]
\centerline{\includegraphics[scale=0.8]{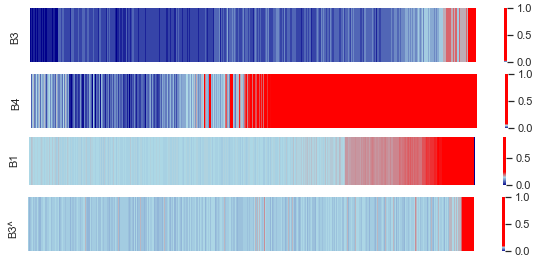}}
\caption{Visualization of anomaly scores from the run-to-failure experiments, where blue color indicates that the system is less likely to be in a faulty state while an abnormal behavior is indicated by the red color bars.}
\label{fig:ims_heatmap}
\end{figure}

A low anomaly score corresponds to healthy behavior while an upward trend (high anomaly score) highlights the abnormal behavior of the machine. Arrival and propagation of fault of for every failing bearing is shown in Fig.~\ref{fig:cm_ims}.  It is clearly visible that the proposed approach is able to identify the initiation of the faulty trend as well as the increasing effect of the deterioration for all four bearings. Fig.~\ref{fig:ims_heatmap} visualizes the anomaly scores and visually indicates the health status of the four faulty bearings S1-B3, S1-B4, and S2-B1 and S3-B3 in parts B3, B4, B1, and B3, respectively. The color bar represents the anomaly score, from 0 (blue) to 1 (red).


\begin{table}[t]
\caption{Model trained on combined healthy data from four RMs is able to detect anomalies without raising any false alarms across all four RMs.}
\begin{center}
\begin{tabular}{|c|c|c|c|c|c|}
\hline
\textbf{Machine} & \textbf{Precision}& \textbf{Recall}& \textbf{TPR} & \textbf{FPR} & \textbf{F1-score}  \\
        \hline
         RM-2 & 1.0 & 1.0 &  1.0 & 0.0 & 1.0 \\
         \hline
         RM-3 & 0.997 & 0.997 &  0.75 & 0.0 & 0.997\\ 
         \hline
         RM-4 & 1.0 & 1.0 &  1.0 & 0.0 & 1.0 \\ 
         \hline
         RM-5 & 1.0 & 1.0 &  1.0 & 0.0 & 1.0 \\
        \hline

\end{tabular}
\label{tab:ps_combined}
\end{center}
\end{table}

\begin{figure*}[t]
\centerline{\includegraphics[scale=0.55]{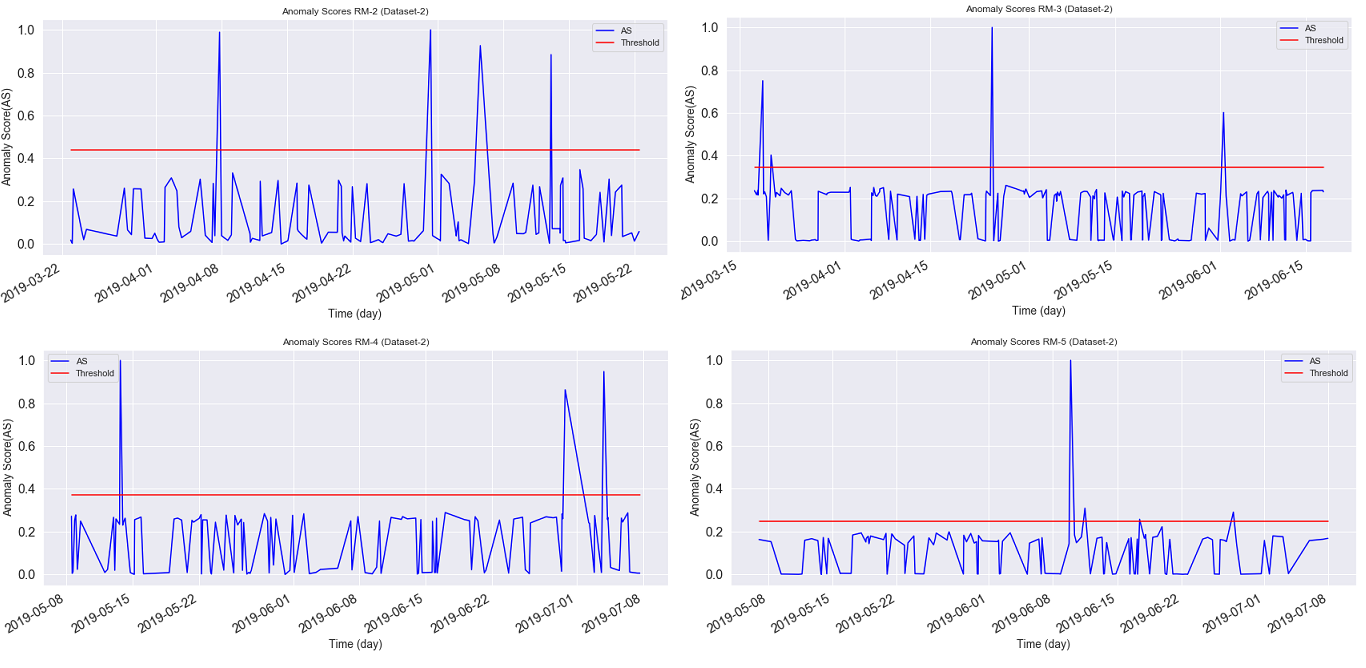}}
\caption{Anomaly detection results obtained from mixing the normal signals from four machines to train the LSTM-AE model and then using the validation data from each machine to calculate the threshold to detect the anomalous samples. All identified anomalies except for one false negative in case of RM-3 were identified correctly by the model.} 
\label{fig:rm2_rm5_result}
\end{figure*}

The same setting was applied to our second dataset where the results were obtained by combining the normal data points from four RMs (RM-2 to RM-5). We divided each RM’s data into three sets: training ${\mathit{Train_{Pi}}}$ (70\%), validation ${\mathit{V_{Pi}}}$ (10\%) and test ${\mathit{Test_{Pi}}}$ (20\%). Train set T was created by combining the four ${{Train_{Pi}}}$ and was used to train the model. The validation sets ${\mathit{V_{Pi}}}$ were used to calculate the threshold ${\tau}$ using the anomaly scores for each RM ${\mathit{P{i}}}$. The final anomaly detection model trained on T was then evaluated on each ${\mathit{Test_{Pi}}}$ set and the obtained results are presented in Table~\ref{tab:ps_combined}, in which the model performs very well in detecting anomalies across four different machines without raising any false alarms. It is worthwhile to notice that combining the training data from multiple RMs increases the performance of the model compared to the performance of the model on individual RMs. The trained LSTM-AE is able to model the healthy working behavior of the RMs and is sensitive to anomalies, as it can be seen in Fig.~\ref{fig:rm2_rm5_result} which shows that the identified anomalies match well with the ground truth. The model can reconstruct the normal signals very well, indicated by a low anomaly score, while it fails to do so whenever it encounters anomalies. 

\subsection{Setting 3: Feature Extraction}
We employ the same LSTM-based autoencoder model to automatically extract time-domain features from vibration signals. The automatically extracted features are compared with a set of manually crafted features, by training an Isolation Forest on both of these two sets and monitoring the performance of the model.

Results obtained from the IMS bearing dataset using the proposed method are compared with the IF-based anomaly detection trained only on manually extracted features. The comparison of results between two Isolation Forest based models is shown in Table~\ref{tab:ims_feature_comparison}, which shows the total number of anomalies detected by the models in each bearing. It’s evident that the S1-B1, S1-B4, S2-B1, and S3-B3 are having comparatively more anomalous samples than the rest of the bearings as these four bearings result in failure according to the ground truth provided. The comparison illustrates that the latent features extracted by the LSTM-AE perform better than the manually engineered features demonstrating the effectiveness of the proposed approach in providing a robust set of features that boost the performance of the anomaly detection model. This can be confirmed by looking at the number of anomalies detected by each model;  for each failing bearing, the IF model based on automatically extracted features detects a greater number of anomalies than the model using the handcrafted set of features. One interpretation of this could be that LSTM-AE provides a set of features that are more sensitive and thus can detect the subtle changes in the vibrations when the fault appears initially and starts developing gradually until the failure finally happens and bearing stops functioning. While on the other hand manually handcrafted features only start labeling signals as anomalous when the fault has fully developed already, and vibrations deviate significantly from the normal signals. 

\begin{table}[htbp]
\caption{Number of anomalies detected by Isolation Forest using automatically extracted features (A\_FE) and manually extracted features (M\_FE): Failing bearings are marked in red and results of the best performing model on these bearings is highlighted with bold entries.}
\begin{center}
\begin{tabular}{|c|c|c|}
\hline
\textbf{Bearing} & \textbf{No. of Anomalies (A\_FE)}& \textbf{No. of Anomalies (M\_FE)}  \\
        \hline
         S1-B1 & 04 & 06  \\
         \hline
         S1-B2 & 07 & 05  \\
         \hline
         \textcolor{red}{S1-B3} & \textbf{39} & 26  \\
         \hline
         \textcolor{red}{S1-B4} & \textbf{441} & 213  \\
         \hline
         
         \textcolor{red}{S2-B1} & \textbf{269} & 133  \\
         \hline
         S2-B2 & 09 & 14  \\
         \hline
         S2-B3 & 07 & 16  \\
         \hline
         S2-B4 & 13 & 11  \\
         \hline
         
         S3-B1 & 24 & 57  \\
         \hline
         S3-B2 & 41 & 35  \\
         \hline
         \textcolor{red}{S3-B3} & \textbf{1016} & 693  \\
         \hline
         S3-B4 & 39 & 51  \\
        \hline

\end{tabular}
\label{tab:ims_feature_comparison}
\end{center}
\end{table}


Multi-step anomaly detection results, where the first step consists of automated features extraction using LSTM-AE and in the second step these features are utilized to train an anomaly detection model (Isolation Forest), for the second dataset are summarized in Table~\ref{tab:p_feature_comparison}. It is important to mention that each RM except for RM-1 in dataset-2 contains very few anomalies (see Table~\ref{tab:rm_2}). IF based on features provided by LSTM-AE (AFE$\_$IF) outperforms IF trained using the manually engineered features (MFE$\_$IF) for all RMs in terms of precision. Although, for RM-3 and RM-5 MFE$\_$IF performed better in terms of TPR. However, it is worthwhile to notice that the model based on automatically engineered features results in less false positives in comparison to the model trained on handcrafted features which had significantly more false positives. As previously stated, the studies that have used handcrafted features obviously are not able to carry the complete system health representation and may not represent the characteristics of the underlying vibration signal under all circumstances. As seen from the results, fault/anomaly detection performance of the conventional methods such as Isolation Forest depends highly on the carefully crafted features. Consequently, this limits the general applicability of these methods.

\begin{table}[ht]
	\caption{Comparison of Isolation Forest trained on automatically extracted features (AFE\_If) and the same model trained using the handcrafted features (MFE\_IF) using the evaluation metrics, row-wise for each RM in Dataset-2.}
	\begin{center}
	\begin{tabular}{|l|l|l|l|l|l|l|}
    \hline
      & \multicolumn{3}{c |}{AFE\_IF}  & \multicolumn{3}{c|}{MFE\_IF} \\
     & P & TPR & FPR & P & TPR & FPR \\
    \hline
    RM-1 & 0.972 & 0.943 & 0.01 & 0.91 & 1.0 & 0.19 \\
    \hline
    RM-2 & 1.0 & 1.0 & 0.0 & 0.976 & 1.0  & 0.23 \\
    \hline
    RM-3 & 0.991 & 0.75 & 0.004 & 0.984 & 1.0 & 0.15 \\
    \hline
    RM-4 & 1.0 & 1.0 & 0.0 & 0.983 & 1.0 & 0.09 \\
    \hline
    RM-5 & 0.979 & 0.25 & 0.0 & 0.977 & 1.0 & 0.14 \\
    \hline
  \end{tabular}
  \end{center}
	\label{tab:p_feature_comparison}
\end{table}


The experimental results obtained in all three settings are able to support the earlier claims that LSTM-AE trained only on the normal/healthy signals is not just able to monitor the health condition of an individual RM but knowledge learned can also be effectively used to detect anomalies in similar yet different RMs. Additionally, it is also shown that latent representations obtained from LSTM-AE can be used as an alternative to manual feature extraction, which requires prior domain and signal processing knowledge.


\section{Conclusion and Future Work}
\label{conclusion}
In conclusion, we have demonstrated an unsupervised method for automated feature extraction from raw vibration signals that can be used for detecting faults and anomalies in rotating machines. A typical condition monitoring system requires feature extraction and decision about the health of the system. The feature extraction component of such systems involves the implementation of signal processing methods for preprocessing the data before using it for anomaly detection. The proposed method fuses the features extraction and anomaly detection modules within one condition monitoring system. Encoding the time-dependent as well as inherent features of the measured vibration time series into a hidden state of an LSTM-based autoencoder enabled the usage of the resulting reconstruction error to be used in an anomaly detection setup. The experimental results on two real-world datasets illustrate the effectiveness of the proposed model and show that LSTM-based autoencoders can extract salient features from vibration signals and achieve high accuracy in fault diagnosis. The model was able to detect the deteriorating condition of all four failing bearings for IMS dataset and achieved an overall F1 score of 99.6\% for the second dataset. As demonstrated in the experiments, the latent representations obtained from the LSMT-AE carry system health information and achieve a higher F1 score than the manually extracted features (97\% and 93\% respectively), indicating a better representation of the characteristics of the vibration signal for the proposed method.

For future work, other alternatives for choosing a threshold value shall be investigated. Also more advanced types of autoencoders such as variational autoencoder can be tried out, which alternatively try to model the data distribution instead of learning to re-create the normal data points, which means datasets with mixed examples (anomalous $\&$ non-anomalous) could also be used for training the model. This paper will help design further advanced feature extraction methods from rotating machines and help build models to monitor the condition of such machines with no or very minimal human supervision. 



\end{document}